\pdfoutput=1

\documentclass[11pt]{article}

\usepackage[final]{acl}

\usepackage{times}
\usepackage{latexsym}

\usepackage[T1]{fontenc}

\usepackage[utf8]{inputenc}

\usepackage{microtype}

\usepackage{inconsolata}

\usepackage{graphicx}
\usepackage{booktabs}

\usepackage{amsmath}
\usepackage{amssymb}
\usepackage{mathtools}
\usepackage{amsthm}

\usepackage{bbm}
\usepackage{multirow}
\usepackage{color, colortbl}

\usepackage{amsmath,amsfonts,bm}

\def\eqref#1{equation~\ref{#1}}

\def\1{\bm{1}}

\def\vz{{\bm{z}}}

\def\mW{{\bm{W}}}

\def\mZ{{\bm{Z}}}

\DeclareMathAlphabet{\mathsfit}{\encodingdefault}{\sfdefault}{m}{sl}
\SetMathAlphabet{\mathsfit}{bold}{\encodingdefault}{\sfdefault}{bx}{n}

\def\gC{{\mathcal{C}}}

\def\gE{{\mathcal{E}}}

\def\gG{{\mathcal{G}}}

\def\gN{{\mathcal{N}}}

\def\gV{{\mathcal{V}}}

\definecolor{Gray}{gray}{0.95}
\definecolor{darkgreen}{RGB}{0, 100, 0}
\definecolor{darkred}{RGB}{139, 0, 0}
\definecolor{First}{HTML}{BFC0FF} 
\definecolor{Second}{HTML}{E7E6FF} 

\title{Can LLMs Convert Graphs to Text-Attributed Graphs?}

\author{
 \textbf{Zehong Wang\textsuperscript{1}},
 \textbf{Sidney Liu\textsuperscript{1}},
 \textbf{Zheyuan Zhang\textsuperscript{1}},
 \textbf{Tianyi Ma\textsuperscript{1}},
\\
 \textbf{Chuxu Zhang\textsuperscript{2}},
 and \textbf{Yanfang Ye\textsuperscript{1}\textsuperscript{*}},
\\
 \textsuperscript{1}University of Notre Dame, Indiana, USA\\
 \textsuperscript{2}University of Connecticut, Connecticut, USA\\
    \texttt{\{zwang43, sliu34, zzhang42, tma2, yye7\}@nd.edu, chuxu.zhang@uconn.edu}
\\
    \textsuperscript{*}Corresponding Author
}

\begin{document}

\maketitle

\begin{abstract}

    Graphs are ubiquitous structures found in numerous real-world applications, such as drug discovery, recommender systems, and social network analysis. To model graph-structured data, graph neural networks (GNNs) have become a popular tool. However, existing GNN architectures encounter challenges in cross-graph learning where multiple graphs have different feature spaces. To address this, recent approaches introduce text-attributed graphs (TAGs), where each node is associated with a textual description, which can be projected into a unified feature space using textual encoders. While promising, this method relies heavily on the availability of text-attributed graph data, which is difficult to obtain in practice. To bridge this gap, we propose a novel method named \underline{\textbf{T}}opology-\underline{\textbf{A}}ware \underline{\textbf{N}}ode description \underline{\textbf{S}}ynthesis (TANS), leveraging large language models (LLMs) to convert existing graphs into text-attributed graphs. The key idea is to integrate topological information into LLMs to explain how graph topology influences node semantics. We evaluate our TANS on text-rich, text-limited, and text-free graphs, demonstrating its applicability. Notably, on text-free graphs, our method significantly outperforms existing approaches that manually design node features, showcasing the potential of LLMs for preprocessing graph-structured data in the absence of textual information. The code and data are available at \url{https://github.com/Zehong-Wang/TANS}. 
\end{abstract}

\section{Introduction}

\begin{figure}[!t]
    \centering
    \includegraphics[width=\linewidth]{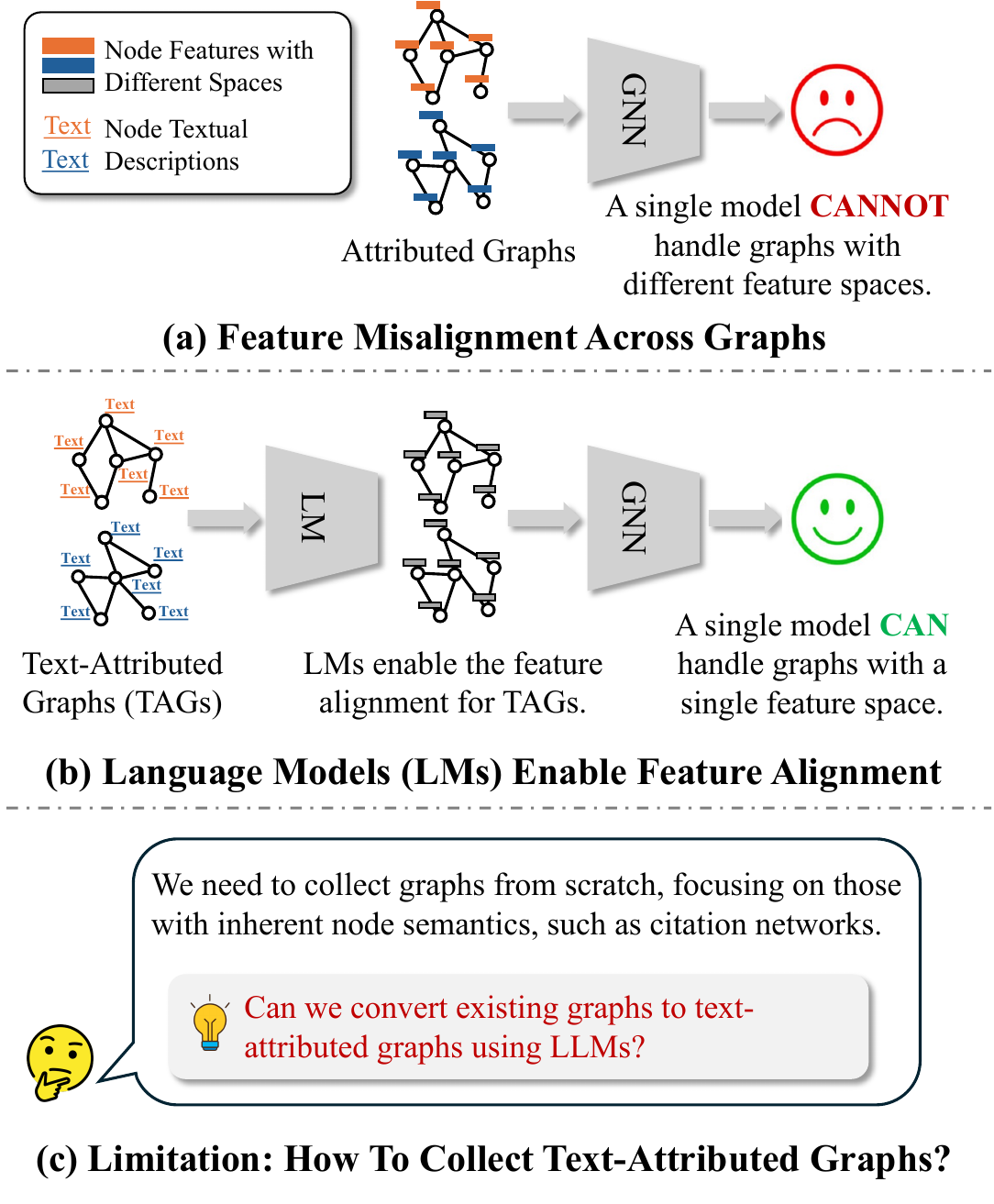}

    \caption{(a) A single GNN model struggles to handle graphs with different feature spaces. (b) Using a textual encoder to align feature spaces across text-attributed graphs (TAGs) facilitates cross-graph learning. (c) However, collecting TAGs is often highly challenging in practice. In this paper, we propose a method to overcome this limitation by automatically generating textual descriptions for nodes in the graph.}

    \label{fig:motivation}
\end{figure}

Graph-structured data are prevalent in many real-world domains, including chemistry \citep{zhao2023gimlet}, social networks \citep{zhang2024diet}, and recommendation systems \citep{zhang2024mopi}. Graph neural networks (GNNs) \citep{kipf2017semisupervised,hamilton2017inductive,velickovic2018graph,gilmer2017neural,wang2023heterogeneous} have emerged as powerful tools for processing such data by learning node representations that capture both the structural and attributed properties of graphs. In each GNN layer, the model first applies a linear transformation to project the input node features or previous layer embeddings into a new space. Then, through message passing \citep{gilmer2017neural}, the model aggregates information from neighboring nodes to update node embeddings.

Although GNNs have shown remarkable success across various applications, a significant limitation is their inability to handle multiple graphs jointly, especially when these graphs have different node feature dimensions \citep{liu2024one,wang2024gft}, as illustrated in Figure \ref{fig:motivation} (a). This is primarily because the input dimension of message-passing GNNs is fixed, making it challenging to accommodate graphs with varying feature spaces. In practice, many real-world graphs possess diverse feature dimensions, preventing a single GNN from being applied across multiple graphs. This limitation poses a substantial barrier to deploying GNNs in scenarios requiring cross-graph learning, such as transfer learning \citep{wang2024tackling}, domain adaptation \citep{dai2022graph}, out-of-distribution detection \citep{li2022ood}, and graph foundation models \citep{wang2024gft}. Overcoming this challenge is crucial for extending the applicability of GNNs to more complex real-world tasks.

To address the challenge of aligning node features across multiple graphs, two main approaches have been proposed. The first approach involves using singular value decomposition (SVD) \citep{yu2024text,zhao2024all} to decompose the original node features of different graphs, aligning the number of dimensions across graphs. However, this method has two limitations: (1) SVD is only applicable to graphs with node features, making it ineffective for graphs that lack such features. (2) While SVD aligns the feature dimensions, it does not ensure that the semantic meaning of the features is consistent across different graphs \citep{yu2024text}. A different approach introduces the concept of text-attributed graphs (TAGs) \citep{yan2023comprehensive}, where each node is associated with a textual description. Researchers use textual encoders \citep{wang2024gft} to transform these descriptions into textual embeddings, better aligning node features across graphs and enabling a single GNN to handle multiple graphs, as shown in Figure \ref{fig:motivation} (b). Compared to SVD, this method provides more meaningful feature alignment and can generalize to unseen TAGs by simply processing the node descriptions \citep{wang2024gft}. However, despite these advantages, TAGs face a significant limitation: the collection of high-quality textual descriptions for all nodes is often impractical, making it challenging to apply this method in real-world scenarios.

In this paper, we address the challenge of collecting textual descriptions for feature alignment in TAGs. Inspired by the success of LLMs in generating synthetic data \citep{tang2023does,long2024llms}, we propose a novel method named \textbf{T}opology-\textbf{A}ware \textbf{N}ode description \textbf{S}ynthesis (TANS) that leverages LLMs to automatically generate node-level textual descriptions for existing graph datasets \citep{tan2024large}, acting as the first work towards this direction. TANS focuses on the node classification task and uses the inherent topological information \citep{bondy2008graph} of each node to guide the LLMs in producing meaningful textual descriptions that capture the role of each node within the graph. Our method is versatile, as it not only applies to graphs without textual descriptions but also enhances graphs with existing rich or limited textual data by generating more informative descriptions. We evaluate our approach on five diverse graph datasets, covering text-rich, text-limited, and text-free scenarios, each with distinct scales and topologies. The experimental results demonstrate the effectiveness of our method, showing superior performance in generating node properties. Moreover, our approach is robust in transfer learning and domain adaptation, further showing the value of LLM-generated descriptions in aligning node features across different graphs.

\begin{figure*}[!t]
    \centering
    \includegraphics[width=\linewidth]{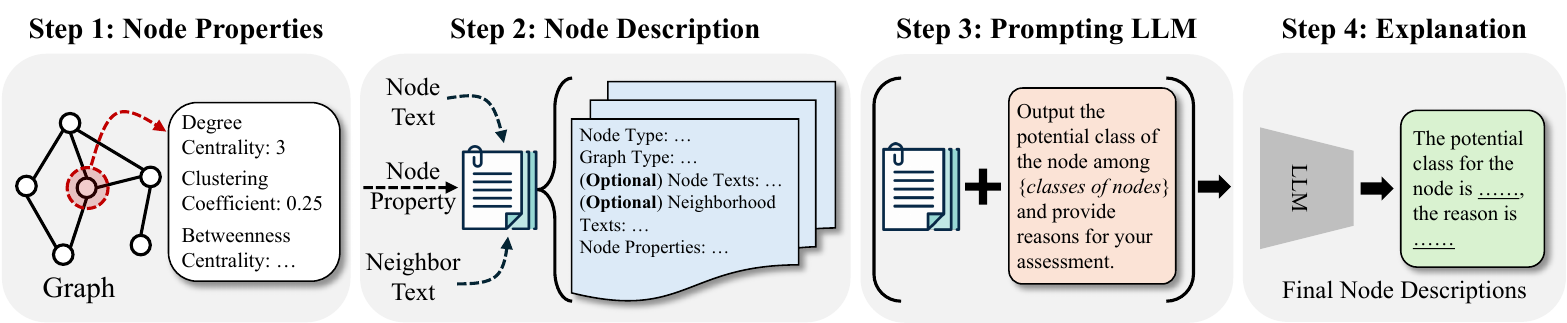}
    \caption{The framework of our topology-aware node description synthesis (TANS).}
    \vspace{-7pt}
    \label{fig:framework}
\end{figure*}

\section{Backgrounds}

\noindent\textbf{Preliminary.} We analyze why existing GNNs struggle to handle multiple graphs with different feature spaces. To understand this limitation, we first review the message-passing process in GNNs. Given a graph $\gG = (\gV, \gE)$ with a node set $\gV$ and edge set $\gE$, where each node $v \in \gV$ is associated with a feature vector $\mathbf{x} \in \mathbb{R}^d$, a GNN encoder $\phi$ takes the graph as input and learns node embeddings $\mZ = \phi(\gV, \gE)$ through message passing. A typical GNN layer is defined as follows:
\begin{equation}
    \nonumber
    \vz_i^{(l)} = \sigma \Bigl(\mW_1 \vz_{i}^{(l-1)} + \frac{\mW_2}{|\gN(i)|} \Bigl(\sum_{j \in \gN(i)} \vz_j^{(l-1)} \Bigr) \Bigr),
\end{equation}
where $\gN(i)$ denotes the neighbors of node $i$, $|\gN(i)|$ is the number of neighbors, $\vz^{(l)}$ is the node embedding at layer $l$, and $\mW_1, \mW_2$ are learnable weight matrices. In the first layer, the model applies a linear transformation to project the input node features, meaning the dimensions of the transformation matrix are fixed based on the input features.

This fixed dimensionality presents a key challenge: when a new graph with different feature dimensions is introduced, the same transformation matrix cannot be used, as it does not accommodate the differing input size. This inherent mismatch is the core reason why a single GNN model cannot effectively handle multiple graphs with different feature spaces.

\noindent\textbf{Related Works.} We present the most relevant works on feature alignment for graphs in this section, with additional references available in Appendix \ref{app:related work}. A common approach is to apply SVD to decompose node features across graphs into a shared feature space \citep{yu2024text,zhao2024all}. However, SVD struggles to ensure fine-grained feature alignment, especially across different graphs. To address this, some methods design advanced GNN models that learn projectors for different graphs \citep{zhao2024all}, but these approaches fail to generalize to unseen graphs \citep{yu2024text}. TAG-based methods \citep{liu2024one,wang2024gft} align features by encoding node textual descriptions using textual encoders. Yet, as noted by \citet{chen2024text}, these embeddings can still reside in different subspaces, limiting their effectiveness. To mitigate this, some works leverage LLMs to generate additional node descriptions for better feature alignment. For instance, TAPE \citep{he2024harnessing} uses LLMs to infer node classes and augment original texts with generated explanations. KEA \citep{chen2024exploring} enhances node embeddings by explaining key terminologies from the original descriptions. However, both methods rely on existing text and cannot handle graphs without textual data.

\section{Method: TANS}

The core idea behind TANS is to leverage topological information as auxiliary knowledge to enhance LLM-generated descriptions for each node. Our method is versatile, applying not only to graphs without textual descriptions but also improving the quality of graphs with existing textual data.

\noindent\textbf{Challenges.} We identify two key challenges in developing TANS: (1) How to identify the most relevant topological information for describing each node? (2) How to effectively integrate LLMs to interpret and utilize this topological information?

\noindent\textbf{Overview.} The framework of TANS is illustrated in Figure \ref{fig:framework}, and it consists of four main steps: (1) Compute topological properties for each node, (2) Use these properties to generate basic node descriptions, (3) Leverage LLMs to predict node roles and explain the reasoning behind these predictions, and (4) Treat the LLM-generated output as the final node descriptions. Before detailing each step, we first present relevant use cases to demonstrate the practicality of our method.

\subsection{Application Scenarios}

\begin{table}[!t]
    \centering
    \resizebox{0.9\linewidth}{!}{
        \begin{tabular}{lccc}
            \toprule
                               & \textbf{TAPE} & \textbf{KEA} & \textbf{TANS} \\ \midrule
            {Text-Rich Graph}  & $\surd$       & $\surd$      & $\surd$       \\ \midrule
            {Text-Limit Graph} & -             & -            & $\surd$       \\ \midrule
            {Text-Free Graph}  & -             & -            & $\surd$       \\
            \bottomrule
        \end{tabular}
    }
    \vspace{-7pt}
    \caption{The comparison between our TANS and two most relevant baselines.}
    \label{tab:compare}
\end{table}

\noindent We define three types of graphs based on the amount of textual information associated with each node: text-rich, text-limit, and text-free graphs. TANS is designed to handle all of these scenarios, while the most relevant baselines, TAPE and KEA, are limited to specific cases, as shown in Table \ref{tab:compare}.

\noindent\textbf{Text-Rich Graphs.} Each node contains abundant textual descriptions that provide sufficient information for downstream tasks. While TAPE and KEA perform well in this scenario, TANS further enhances the node descriptions by incorporating topological information.

\noindent\textbf{Text-Limit Graphs.} Nodes have only sparse textual descriptions, which may lack sufficient detail for downstream tasks. TANS supplements this limited information with topological knowledge, making it more effective than baselines.

\noindent\textbf{Text-Free Graphs.} No textual descriptions are available for nodes, leaving topological information as the only resource for downstream tasks. TANS excels in this scenario, where other methods are not applicable.

\subsection{Step 1: Graph Properties}

To balance effectiveness in describing node-level characteristics and computational efficiency, we select the following five graph properties.

\noindent\textbf{Degree Centrality.} This property measures the number of directly connected nodes for a target node, capturing its localized importance or influence \citep{zhang2017degree}. It helps LLMs determine whether a node is central or peripheral within the graph:
\begin{equation}
    \nonumber \gC_D (v) = \deg(v) = | \gN(v) |,
\end{equation}
where $\deg$ indicates the node degree.

\noindent\textbf{Betweenness Centrality.} This property measures how frequently a node lies on the shortest paths between other nodes, highlighting its role in facilitating communication or information flow within the graph \citep{zhang2017degree}. This helps LLMs to identify nodes that act as key intermediaries for generating more informative descriptions:
\begin{equation}
    \nonumber \gC_B (v) = \sum_{s \neq v \neq t \in \gV} \frac{\sigma_{st}(v)}{\sigma_{st}},
\end{equation}
where $\sigma_{st}$ is the total number of shortest paths from node $s$ to node $t$, and $\sigma_{st}(v)$ is the number of those paths that pass through $v$ (excluding endpoints).

\noindent\textbf{Closeness Centrality.} This property measures how close a node is to all other nodes in the graph by calculating the average distance from a given node to every other node. It reflects the node's global centrality and how efficiently information can spread from it across the graph \citep{zhang2017degree}. Thus, it helps LLMs to capture the global influence of nodes:
\begin{equation}
    \nonumber \gC_C (v) = \frac{N-1}{\sum_{v \in \gV, u \neq v} d(u, v)},
\end{equation}
where $N$ is the number of nodes, and $d(u,v)$ is the shortest distance between nodes $u$ and $v$.

\noindent\textbf{Clustering Coefficient.} This property measures the likelihood that a node's neighbors are also connected to each other, indicating the formation of triangle-like structures in the graph. It provides LLMs with insights into the local transitivity of the network \citep{saramaki2007generalizations}, which is crucial for understanding the cohesiveness of a node's neighborhood:
\begin{equation}
    \nonumber \gC_{tri} = \frac{2 T(v)}{\deg(v) (\deg(v) - 1)},
\end{equation}
where $T(v)$ is the number of triangles that include node $v$, and $\deg(v) (\deg(v) - 1)$ represents the maximum possible number of triangles around node $v$. A value of $\gC_{tri} = 1$ indicates that all of neighboring nodes are fully connected, while $\gC_{tri} = 0$ suggests that the node is isolated in its neighborhood.

\noindent\textbf{Square Clustering Coefficient.} Similar to the clustering coefficient, this property measures the tendency of nodes to form square-like structures rather than triangle-like ones. By capturing more complex interactions among nodes, it provides LLMs with a deeper understanding of the correlations within a node's neighborhood \citep{zhang2008clustering}.

\subsection{Step 2: Generate Basic Node Descriptions}

\begin{table*}[!t]
    \centering
    \resizebox{\linewidth}{!}{
        \begin{tabular}{p{4.1cm} p{15cm}}
            \toprule
            \rowcolor{Gray}   \multicolumn{2}{l}{\textit{{Step 2: Generate Basic Node Descriptions}}}                                                                                                                                         \\ \midrule
            \textbf{Prefix}                   & {\tt Given a node from a \{Graph Type\} graph, where the node type is \{Node Type\} with \{Node Number\} nodes, and the edge type is \{Edge Type\} with \{Edge Number\} edges.}               \\ \midrule
            \textbf{Node Text (Optional)}     & {\tt The original node description is \{\textit{Original Textual Descriptions}\}.}                                                                                                            \\ \midrule
            \textbf{Neighbor Text (Optional)} & {\tt The following are the textual information of \{$k$\} connected nodes. The descriptions are: \{\textit{Textual Descriptions of Selected Neighborhoods}\}. }                               \\ \midrule
            \textbf{Node Property}            & {\tt The value of \{\textit{Node Property}\} is \{Value of The Given Property\}, ranked as \{Rank of The Node\}\% among \{\textit{Node Number}\} nodes.}                                      \\ \midrule\midrule
            \rowcolor{Gray}   \multicolumn{2}{l}{\textit{{Step 3: Prompting LLMs}}}                                                                                                                                                           \\ \midrule
            \textbf{Suffix}                   & {\tt Output the potential \{$k$\} classes of the node and provide reasons for your assessment. The classes include \{\textit{Classes of Nodes}\}. Your answer should be less than 200 words.} \\
            \bottomrule
        \end{tabular}
    }
    \vspace{-7pt}
    \caption{Prompt templates. }
    \label{tab:template}
\end{table*}

We generate basic node descriptions using the computed node properties, which are then fed into LLMs for inference. These descriptions are composed of four components, as shown in Table \ref{tab:template}.

\noindent\textbf{Prompt 1: Prefix.} This part provides basic information about the graph, including its type and the type of nodes, helping LLMs detect key properties and interpret the following content more effectively. For example, in citation graphs, the LLMs might prioritize textual descriptions since they provide rich information about papers \citep{he2024harnessing}. In contrast, in social networks, topological features like the clustering coefficient or degree centrality may be more relevant \citep{zhang2017degree}, reflecting close friendships or node popularity, respectively.

\noindent\textbf{Prompt 2: (Optional) Node Text.} This component incorporates the original node textual descriptions (if applicable), enabling the method to handle graphs that have inherent textual data. It also helps capture neighborhood information more effectively when describing the target node.

\noindent\textbf{Prompt 3: (Optional) Neighbor Text.} This component stores the textual descriptions of neighboring nodes. We randomly select $k = 5$ neighbors to provide additional context. This is especially important for text-limited graphs, where original descriptions may be insufficient, and 1-hop neighborhood information has proven to be informative \citep{han2023mlpinit,ju2024does}, as supported by our experimental results in Table \ref{tab:ablation}.

\noindent\textbf{Prompt 4: Node Property.} This component appends the pre-processed node properties to the prompt. Unlike methods that input the entire graph for inference \citep{guo2023gpt4graph,wang2024can}, our approach explicitly injects topological knowledge into the LLMs, making them more controllable. We also provide the ranking of nodes based on these properties, ensuring that the LLMs better understand their relative importance.

\subsection{Step 3: Prompting LLMs}




The basic node descriptions from the previous step are fed into an LLM for inference. To ensure robustness and transferability, we aim to generate descriptions that are consistent across different graphs, so that the resulting textual embeddings remain close in the feature space. In our experiment, we use the public GPT-4o-mini interface for prompting.

\noindent\textbf{Prompt 5: Suffix.} To achieve this, we avoid outputting overly specific or uninterpretable knowledge. Instead, we aim for general descriptions by providing the potential node classes and having the LLM analyze the correlation between the basic descriptions and these classes, generating the top-$k$ predictions along with corresponding explanations. The specific prompt format is shown in Table \ref{tab:template}. If the number of classes exceeds 3, we set $k = 3$; otherwise, $k$ is set to 1.

\subsection{Step 4: Explanations}


We use the LLM-generated output as the final node descriptions, which explain why a node likely belongs to certain classes. For text-rich and text-limited graphs, we append the generated descriptions to the original text and then use a textual encoder to produce the node embeddings. For text-free graphs, the generated text serves as the node description, and we similarly apply a textual encoder for embedding.

\section{Experiments}

\subsection{Experimental Setup}


\begin{table}[!t]
    \centering
    \resizebox{\linewidth}{!}{
        \begin{tabular}{lrrrrl}
            \toprule
                            & \textbf{Nodes} & \textbf{Edges} & \textbf{Classes} & \textbf{Graph Types} \\ \midrule
            \texttt{Cora}   & 2,708          & 10,556         & 7                & Text-rich / -limit   \\
            \texttt{Pubmed} & 19.717         & 88,648         & 3                & Text-rich / -limit   \\ \midrule
            \texttt{USA}    & 1,190          & 28,388         & 4                & Text-free            \\
            \texttt{Europe} & 399            & 12,385         & 4                & Text-free            \\
            \texttt{Brazil} & 131            & 2,137          & 4                & Text-free            \\ \bottomrule
        \end{tabular}
    }
    \vspace{-7pt}
    \caption{The statistics of datasets.
    }
    \label{tab:data}
\end{table}

\begin{table*}[!t]
    \resizebox{\linewidth}{!}{
        \begin{tabular}{clcccccccc}
            \toprule
                                                        &              & \multicolumn{4}{c}{\bf Low-Label} & \multicolumn{4}{c}{\bf High-Label}                                                                                                                                     \\ \cmidrule(lr){3-6} \cmidrule(lr){7-10}
                                                        &              & \textbf{GCN}                      & \textbf{GAT}                       & \textbf{MLP}            & \textbf{Avg.}  & \textbf{GCN}          & \textbf{GAT}          & \textbf{MLP}          & \textbf{Avg.}  \\ \midrule
            \multirow{6}{*}{\rotatebox{90}{\tt Cora}}   & Raw Feat.    & 78.39 ± 1.69                      & 79.31 ± 1.70                       & 66.18 ± 4.95            & 74.63          & 83.10 ± 1.69          & 82.45 ± 1.23          & 64.56 ± 1.95          & 76.70          \\ \cmidrule{2-10}
                                                        & Raw Text     & 79.19 ± 1.63                      & 80.09 ± 1.57                       & 70.55 ± 1.40            & 76.61          & 87.45 ± 1.15          & 85.72 ± 1.47          & 78.95 ± 1.45          & 84.04          \\
                                                        & \quad + TAPE & 79.64 ± 1.36                      & {80.28 ± 1.37}                     & {70.97 ± 2.02}          & {76.96}        & 87.69 ± 1.34          & 86.21 ± 1.33          & {80.07 ± 1.72}        & 84.66          \\
                                                        & \quad + KEA  & {80.08 ± 1.71}                    & 79.80 ± 1.58                       & 70.72 ± 1.51            & 76.87          & {87.94 ± 1.28}        & {86.58 ± 1.10}        & 79.90 ± 1.83          & {84.81}        \\ \cmidrule{2-10}
            \rowcolor{Gray}                             & \quad + TANS & \textbf{81.26 ± 1.48}             & \textbf{81.08 ± 1.62}              & \textbf{72.47   ± 1.96} & \textbf{78.27} & \textbf{88.91 ± 1.57} & \textbf{88.23 ± 1.26} & \textbf{81.44 ± 1.42} & \textbf{86.19} \\ \midrule\midrule
            \multirow{6}{*}{\rotatebox{90}{\tt Pubmed}} & Raw Feat.    & 75.39 ± 1.51                      & 74.59 ± 1.36                       & 68.01 ± 1.99            & 72.66          & 84.10 ± 0.55          & 84.31 ± 0.66          & 80.56 ± 0.30          & 82.99          \\  \cmidrule{2-10}
                                                        & Raw Text     & {76.97 ± 1.95}                    & 75.50 ± 2.03                       & 70.78 ± 2.00            & 74.42          & 87.49 ± 0.54          & 87.20 ± 0.51          & 82.58 ± 0.38          & 85.76          \\
                                                        & \quad + TAPE & 76.50 ± 3.27                      & 75.30 ± 1.92                       & 71.06 ± 2.13            & 74.29          & {88.21 ± 0.62}        & {87.80 ± 0.48}        & 83.98 ± 0.59          & 86.66          \\
                                                        & \quad + KEA  & 76.88 ± 1.73                      & {75.74 ± 2.06}                     & {71.32 ± 2.51}          & {74.65}        & 88.10 ± 0.49          & 87.77 ± 0.50          & {85.33 ± 0.41}        & {87.07}        \\   \cmidrule{2-10}
            \rowcolor{Gray}                             & \quad + TANS & \textbf{78.01 ± 2.26}             & \textbf{76.99 ± 2.02}              & \textbf{73.64 ± 2.59}   & \textbf{76.21} & \textbf{88.96 ± 0.39} & \textbf{87.98 ± 0.48} & \textbf{88.84 ± 0.43} & \textbf{88.59} \\
            \bottomrule
        \end{tabular}
    }
    \vspace{-7pt}
    \caption{Results on text-rich citation graphs.}
    \vspace{-5pt}
    \label{tab:text-rich}
\end{table*}

\begin{table*}[!t]
    \resizebox{\linewidth}{!}{
        \begin{tabular}{lcccccccc}
            \toprule
                                 & \multicolumn{4}{c}{\bf Low-Label} & \multicolumn{4}{c}{\bf High-Label}                                                                                                                                     \\ \cmidrule(lr){2-5}\cmidrule(lr){6-9}
                                 & \texttt{Europe}                   & \texttt{USA}                       & \texttt{Brazil}        & \textbf{Avg.}  & \texttt{Europe}       & \texttt{USA}          & \texttt{Brazil}        & \textbf{Avg.}  \\ \midrule
            Raw Feat. (One-Hot)  & 51.89 ± 2.75                      & 52.74 ± 2.25                       & 65.15 ± 15.93          & 56.59          & 54.61 ± 5.91          & 60.88 ± 3.83          & 49.88 ± 11.50          & 55.12          \\ \midrule
            Node Degree          & 54.69 ± 3.35                      & {59.93 ± 2.21}                     & {71.82 ± 12.28}        & {62.15}        & 55.72 ± 5.12          & {64.36 ± 3.18}        & 63.83 ± 9.35           & 61.30          \\
            Eigenvector          & 55.80 ± 2.47                      & 57.72 ± 2.19                       & 62.42 ± 13.83          & 58.65          & \textbf{58.15 ± 4.51} & 63.66 ± 2.88          & 65.06 ± 8.95           & 62.29          \\
            Random Walk          & {56.70 ± 2.47}                    & 56.11 ± 2.11                       & 69.70 ± 14.34          & 60.84          & 55.71 ± 4.01          & 62.80 ± 3.01          & {68.40 ± 9.65}         & {62.30}        \\ \midrule
            \rowcolor{Gray} TANS & \textbf{56.87 ± 3.14}             & \textbf{61.08 ± 2.71}              & \textbf{80.61 ± 12.14} & \textbf{66.19} & {56.05 ± 5.48}        & \textbf{65.32 ± 3.16} & \textbf{71.60 ± 10.66} & \textbf{64.32} \\
            \bottomrule
        \end{tabular}
    }
    \vspace{-7pt}
    \caption{Results on text-free airport graphs with GCN backbone. }
    \vspace{-10pt}
    \label{tab:text-free gcn}
\end{table*}

\noindent\textbf{Dataset.} We use five graph-structured datasets in our experiments: \texttt{Cora}, \texttt{Pubmed}, \texttt{USA}, \texttt{Europe}, and \texttt{Brazil}, with statistics provided in Table \ref{tab:data}. \texttt{Cora} and \texttt{Pubmed} are citation networks \citep{he2024harnessing}, where nodes represent papers and edges represent citations. Each node contains paper titles and abstracts, and the classes correspond to paper types. These graphs can be either text-rich (using both titles and abstracts) or text-limited (using only titles or abstracts). \texttt{USA}, \texttt{Europe}, and \texttt{Brazil} are text-free airport datasets \citep{ribeiro2017struc2vec}, where nodes represent airports and edges represent flight connections. The goal is to classify airports based on their activity levels.


\noindent\textbf{Baselines.} We compare several feature alignment methods. For graphs with textual descriptions, the primary baselines are \textbf{TAPE} \citep{he2024harnessing} and \textbf{KEA} \citep{chen2024exploring}, as discussed in related works. In our experiments, we append the generated texts to the original node descriptions rather than using their original, more complex training paradigms, as our focus is on aligning feature spaces across graphs for cross-graph learning. We also evaluate models that rely solely on original textual descriptions or node features. For text-free graphs, where TAPE and KEA are not applicable, we compare against methods that generate node features from graph topologies, such as Node Degree \citep{ribeiro2017struc2vec}, Eigenvector \citep{dwivedi2023benchmarking}, and Random Walk \citep{dwivedi2022graph}. For these methods, we set the number of feature dimensions to 32, which we found empirically to provide good performance.

\noindent\textbf{Evaluation Protocol.} We run each experiment 30 times with different random seeds to reduce the impact of randomness. The node classification results are reported on the test set, using the model that performs best on the validation set. We use accuracy as the evaluation metric and employ GCN \citep{kipf2017semisupervised}, GAT \citep{velickovic2018graph}, and MLP as backbone models. For the textual encoder, we follow \citep{chen2024text} to use MiniLM \citep{wang2020minilm} otherwise specifically indicated. The hyper-parameters are presented in Appendix \ref{app:exp hyper}.

\begin{table*}[!t]
    \resizebox{\linewidth}{!}{
        \begin{tabular}{lccccccc}
            \toprule
            \multicolumn{1}{r}{Source $\to$} & \multicolumn{2}{c}{\tt USA} & \multicolumn{2}{c}{\tt Europe} & \multicolumn{2}{c}{\tt Brazil} &                                                                                        \\ \cmidrule(lr){2-3} \cmidrule(lr){4-5} \cmidrule(lr){6-7}
            \multicolumn{1}{r}{Target $\to$} & \texttt{Europe}             & \texttt{Brazil}                & \texttt{USA}                   & \texttt{Brazil}       & \texttt{USA}          & \texttt{Europe}       & \textbf{Avg.}  \\ \midrule
            Raw Feat. (One-Hot)              & -                           & -                              & -                              & -                     & -                     & -                     & -              \\
            \quad + SVD                      & 30.55 ± 4.61                & 34.23 ± 5.19                   & 45.90 ± 3.90                   & 57.21 ± 5.30          & 24.95 ± 3.19          & 45.48 ± 2.58          & 39.72          \\ \midrule
            Node Degree                      & 46.61 ± 1.54                & 52.29 ± 3.91                   & \textbf{53.40 ± 1.09}          & {66.76 ± 3.85}        & {54.35 ± 2.22}        & 51.85 ± 2.14          & {54.21}        \\
            Eigenvector                      & 37.73 ± 3.08                & 32.79 ± 4.49                   & 50.12 ± 1.76                   & 61.49 ± 4.33          & 25.43 ± 0.98          & 50.96 ± 4.42          & 43.09          \\
            Random Walk                      & {48.79 ± 2.60}              & {58.13 ± 3.38}                 & 49.45 ± 1.59                   & 62.38 ± 5.98          & 44.82 ± 1.65          & {52.71 ± 2.09}        & 52.71          \\ \midrule
            \rowcolor{Gray}  TANS            & \textbf{50.99 ± 3.31}       & \textbf{67.17 ± 4.68}          & {51.88 ± 2.82}                 & \textbf{71.59 ± 3.97} & \textbf{54.96 ± 1.80} & \textbf{53.79 ± 2.15} & \textbf{58.40} \\
            \bottomrule
        \end{tabular}
    }
    \vspace{-7pt}
    \caption{Domain adaptation results on text-free airport graphs. }
    \vspace{-10pt}
    \label{tab:da}
\end{table*}

\begin{table}[!t]
    \resizebox{\linewidth}{!}{
        \begin{tabular}{ccccc}
            \toprule
            \multicolumn{2}{l}{Node Text $\to$}         & Title         & Abstract     & Title + TANS                         \\ \midrule
            \multirow{4}{*}{\rotatebox{90}{\tt Cora}}   & \textbf{GCN}  & 79.06 ± 1.68 & 77.89 ± 2.26 & \textbf{79.94 ± 1.62} \\
                                                        & \textbf{GAT}  & 77.89 ± 1.64 & 56.59 ± 8.49 & \textbf{80.34 ± 1.25} \\
                                                        & \textbf{MLP}  & 57.42 ± 2.46 & 49.15 ± 3.92 & \textbf{68.35 ± 1.85} \\ \cmidrule{2-5}
                                                        & \textbf{Avg.} & 71.46        & 61.21        & \textbf{76.21}        \\ \midrule\midrule
            \multirow{4}{*}{\rotatebox{90}{\tt Pubmed}} & \textbf{GCN}  & 75.17 ± 2.09 & 77.20 ± 2.04 & \textbf{81.40 ± 2.03} \\
                                                        & \textbf{GAT}  & 74.43 ± 2.46 & 76.81 ± 2.51 & \textbf{80.03 ± 1.68} \\
                                                        & \textbf{MLP}  & 66.31 ± 2.73 & 72.26 ± 2.42 & \textbf{76.33 ± 2.65} \\ \cmidrule{2-5}
                                                        & \textbf{Avg.} & 71.97        & 75.42        & \textbf{79.25}        \\
            \bottomrule
        \end{tabular}
    }
    \vspace{-7pt}
    \caption{Results on text-limit citation graphs.}
    \vspace{-10pt}
    \label{tab:text-limit}
\end{table}

\subsection{Single-Graph Learning}

\noindent\textbf{Setting.} We evaluate model performance by training from scratch, following the settings from \citep{chen2024exploring}. Two evaluation settings are used: low-labeling and high-labeling. In the low-label setting, we randomly select 20 nodes per class for training, 30 nodes per class for validation, and use the remaining nodes for testing. For the smaller \texttt{Brazil} dataset, we use 10 nodes per class for training and 20 for validation. In the high-label setting, we randomly split the nodes into 60\%/20\%/20\% for training, validation, and testing.

\noindent\textbf{Text-Rich Graphs.} Table \ref{tab:text-rich} presents the performance of our models on \texttt{Cora} and \texttt{Pubmed} under low-label and high-label settings, using GCN, GAT, and MLP backbones. The results show that node features generated by advanced textual encoders outperform the original features. While methods like KEA and TAPE improve performance with additional textual information, our approach achieves superior results. This is likely due to the incorporation of graph topological properties, which provide a deeper understanding of node roles within the graph. Notably, our method shows the largest improvement with MLPs, demonstrating the advantage of using topological information to enhance node feature quality.



\noindent\textbf{Text-Limit Graphs.} Table \ref{tab:text-limit} shows the performance of our models in the text-limit setting using the GCN backbone under low-label conditions. Since TAPE and KEA cannot be applied here, we compare the performance of (1) using only titles, (2) using only abstracts, and (3) using titles combined with our method. As expected, the performance in the text-limit setting is lower than in the text-rich setting. However, our method improves performance by about 5\% on average compared to the best baselines, highlighting the effectiveness of incorporating topological properties to enhance node descriptions in graphs with limited text.


\noindent\textbf{Text-Free Graphs.} We report model performance on \texttt{USA}, \texttt{Europe}, and \texttt{Brazil} using the GCN backbone in Table \ref{tab:text-free gcn} and the MLP backbone in Table \ref{tab:text-free mlp}. Our method significantly outperforms existing approaches that use graph properties to generate node features. We attribute this improvement to two factors: (1) Our method utilizes a larger set of graph topological properties, which better describe node characteristics, and (2) LLMs analyze the relationship between the prompts and potential classes, where LLM's inherent knowledge helps infer node classes based on the provided textual descriptions, offering additional information. These results highlight the potential of our method to generate effective node descriptions even for graphs without initial textual data, enabling a unified model to process multiple graphs.

\subsection{Cross-Graph Learning}


\noindent\textbf{Setting.} We consider two cross-graph learning settings: domain adaptation and pretrain \& finetune. In domain adaptation, we train the model on a source graph and evaluate its performance on a target graph, with 20\% of the data used for validation and 80\% for testing. For pretrain \& finetune, we pretrain the model on the source graph and fine-tune it on the target graph using the high-label setting described in the previous section. Additional details for both settings are provided in Appendix \ref{app:exp da} and Appendix \ref{app:exp pt}, respectively.

\begin{table}[!t]
    \resizebox{\linewidth}{!}{
        \begin{tabular}{lcc}
            \toprule
                                           & \texttt{Cora} $\to$ \texttt{Pubmed} & \texttt{Pubmed} $\to$ \texttt{Cora} \\ \midrule
            Raw Feat.                      & -                                   & -                                   \\
            \quad + SVD                    & 70.39 ± 6.12                        & 70.48 ± 3.71                        \\ \midrule
            Raw Text                       & 75.77 ± 2.96                        & 79.62 ± 2.04                        \\
            \quad + TAPE                   & 75.60 ± 2.39                        & 79.25 ± 2.06                        \\
            \quad + KEA                    & 75.25 ± 2.50                        & 79.59 ± 1.61                        \\ \midrule
            \rowcolor{Gray}   \quad + TANS & \textbf{76.14 ± 2.28}               & \textbf{80.05 ± 1.74}               \\ \bottomrule
        \end{tabular}
    }
    \vspace{-7pt}
    \caption{Transfer learning results on citation networks.}
    \vspace{-10pt}
    \label{tab:transfer}
\end{table}


\noindent\textbf{Pretrain \& Finetune.} In this setting, we use text-rich \texttt{Cora} and \texttt{Pubmed}, with results shown in Table \ref{tab:transfer}. The table shows that simply applying SVD to the original node features results in significantly lower performance. Additionally, existing methods perform poorly in the transfer learning setting, sometimes even worse than using the original textual descriptions. This may be due to the excess information provided by the generated text, which limits generalization and obscures shared patterns across graphs. In contrast, our TANS incorporates topological information to generate node descriptions, proving more robust in transfer learning. One possible reason is that the topological properties capture shared structures across citation networks, leading to more reliable text generation\footnote{Although we observe negative transfer in some experiments, this is a common issue in graph learning \citep{wang2024tackling} and beyond the scope of this paper, as our focus is on demonstrating the potential of using LLMs for feature alignment in cross-graph training.}.


\noindent\textbf{Domain Adaptation.} The results for domain adaptation across the text-free graphs \texttt{USA}, \texttt{Brazil}, and \texttt{Europe} are presented in Table \ref{tab:da}, using GCN as the backbone. Our method achieves the highest average performance of 58.40, significantly outperforming the second-best result of 54.21. Additionally, it achieves the best performance on 5 out of 6 datasets. These results demonstrate that our method generates reliable textual descriptions for nodes in text-free graphs, outperforming approaches that focus on individual topological properties. The generated texts also exhibit some transferability, highlighting the potential generalization capability of our approach.

\begin{table}[!t]
    \resizebox{\linewidth}{!}{
        \begin{tabular}{lcc}
            \toprule
                                        & \textbf{Text-Rich} & \textbf{Text-Limit} \\ \midrule
            Raw Text                    & 76.61              & 71.46               \\ \midrule
            \quad + TANS w.o. neighbors & 77.54              & 72.92               \\
            \quad + TANS                & \textbf{78.27}     & \textbf{76.21}      \\
            \bottomrule
        \end{tabular}
    }
    \vspace{-7pt}
    \caption{Ablation results without neighborhood information in generating node descriptions. We report the average performance across three backbones on \texttt{Cora} in low-label setting. Full results are provided in Table \ref{tab:ablation full}.}
    \label{tab:ablation}
\end{table}

\begin{table}[!t]
    \centering
    \resizebox{\linewidth}{!}{
        \begin{tabular}{lcccc}
            \toprule
            \textbf{Methods}      & Raw Text & TAPE  & KEA   & TANS           \\ \midrule
            \textbf{Avg. Results} & 80.11    & 80.50 & 80.57 & \textbf{81.22} \\
            \bottomrule
        \end{tabular}
    }
    \vspace{-7pt}
    \caption{Average performance of four textual encoders on the text-rich \texttt{Cora} dataset with low-label setting and GCN backbone. Full results are provided in Table \ref{tab:text enc full}.}
    \label{tab:text enc}
    \vspace{-5pt}
\end{table}

\subsection{Ablation Study}

\noindent\textbf{Prompts in Generating Node Descriptions.} We analyze the impact of different prompt components, specifically focusing on the role of neighboring node textual information. When this information is excluded, only node properties and optional node descriptions are used. We conduct an ablation study by creating a variant that omits the neighborhood textual descriptions. The averaged results across three backbones are shown in Table \ref{tab:ablation}, with full results in Table \ref{tab:ablation full}. The results demonstrate that our method improves performance on text-attributed graphs, whether using both topological properties and neighborhood descriptions or just topological properties. Notably, on text-limited graphs, the inclusion of neighborhood information is crucial, as the model's performance drops from 76.21 to 72.92 when neighborhood descriptions are removed. This highlights the importance of incorporating topological and neighborhood information to enhance node descriptions.


\noindent\textbf{Textual Encoders.} To evaluate the robustness of the generated textual descriptions, we tested four different textual encoders: MiniLM \citep{wang2020minilm}, Albert \citep{Lan2020ALBERT}, Roberta \citep{liu2020roberta}, and MPNet \citep{song2020mpnet}. The average performance on the text-rich \texttt{Cora} dataset with a low-label setting and GCN backbone is shown in Table \ref{tab:text enc}, with full results in Table \ref{tab:text enc full}. Our method consistently achieves the best performance, demonstrating the robustness and high quality of the generated textual descriptions.

\section{Discussion}

\noindent\textbf{Expanding to More Graph-Related Tasks.} In our experiments, the proposed TANS achieves desirable performance on node classification tasks for citation and airport networks, demonstrating the potential of LLMs in understanding node properties based on graph topology. This success motivates us to explore the potential of LLMs in understanding edge and graph properties, extending our method to edge-level and graph-level tasks. We plan to investigate these extensions in future work.


\noindent\textbf{Converting Basic Attributed Graphs.} Our proposed TANS converts existing graphs into text-attributed graphs, facilitating feature alignment in graph preprocessing. Although our focus is on graphs classified by their associated textual descriptions, we believe that attributed graphs, where node features are generated through feature engineering, can also be converted into text-attributed graphs due to the inherent semantics of each feature dimension. For instance, \texttt{Cora} uses a 1433-dimensional one-hot encoding, with each dimension corresponding to a keyword, and \texttt{Pubmed} uses a 500-dimensional TF-IDF vector, where each dimension represents a keyword. By leveraging these inherent semantics, we can convert original node features into textual descriptions. Whether these converted graphs are classified as text-rich or text-limited will depend on the specific case. We plan to explore this conversion process in future work.

\section{Conclusion}

In this work, we explore the ability of LLMs to convert existing graphs to text-attributed graphs by generating node descriptions in graphs, regardless of whether the graphs contain textual information. Our proposed TANS enables LLMs to incorporate graph topological information when generating node descriptions, allowing for the alignment of node features across graphs. Experimental results demonstrate the superiority of our method across text-rich, text-limited, and text-free graphs in training from the scratch, domain adaptation, and transfer learning settings.

\section*{Limitations}

One limitation of our work is the exclusion of large-scale graphs (with more than 100,000 nodes) from our experiments. Applying TANS to such large graphs is expensive, as generating textual descriptions requires querying GPT for each node, which significantly increases time and cost. This limitation also restricted our ability to conduct more extensive ablation studies on prompt design. However, the ablation studies we performed still provide meaningful insights into how our method works, and future work could explore more efficient template designs to further optimize the process.

Additionally, we used GPT-4o-mini for querying, which may have a lower capacity compared to GPT-4o. Despite this, our experimental results were still highly desirable. It remains uncertain whether GPT-4o would significantly outperform GPT-4o-mini, and further investigation into this aspect could be part of future research.

\section*{Ethical Considerations}

Our method serves as a tool for generating textual descriptions for graph-structured data using LLMs. However, there is potential for the generated content to include biased or harmful information. To mitigate this risk, more careful prompt design, including clear instructions and guidelines, can help steer the LLMs toward generating positive and accurate content. Additionally, users must be mindful of ethical concerns such as bias in the data and ensure responsible use of the tool in different applications.

\section*{Acknowledgement}

This work was partially supported by the NSF under grants IIS-2321504, IIS-2334193, IIS-2340346, IIS-2217239, CNS-2426514, CNS-2203261, and CMMI-2146076. Any opinions, findings, and conclusions or recommendations expressed in this material are those of the authors and do not necessarily reflect the views of the sponsors.

\bibliography{custom}

\newpage
\appendix

\section{More Related Works}
\label{app:related work}

\noindent\textbf{Graph Neural Networks.} Graph neural networks (GNNs) \citep{liu2023fair,wang2024select,wang2024training,liu2024can} are effective in various graph learning tasks by utilizing the message passing framework. For example, GCN \citep{kipf2017semisupervised} leverages the Laplacian matrix for message passing, MPNN \citep{gilmer2017neural} formally defines the message passing framework, GraphSAGE \citep{hamilton2017inductive} extends it to inductive learning, and GAT \citep{velickovic2018graph} introduces attention mechanisms. Further works \citep{wang2019heterogeneous,wang2023heterogeneous,zhang2024diet} have extended message passing to various graph types and applications. However, a key limitation of message passing GNNs is their inability to handle graphs with different feature spaces \citep{liu2024one,wang2024gft}, highlighting the need for effective feature alignment methods across graphs.

\noindent\textbf{Manually Designed Node Features.} Another approach involves manually designing node features using topological information. For example, node degrees can be represented using one-hot encoding to describe node properties \citep{ribeiro2017struc2vec}. Additionally, methods such as the eigenvector of the graph Laplacian or random walk-based techniques like node2vec \citep{grover2016node2vec} can be used to generate node embeddings based solely on topological properties.

\section{Experimental Setting}
\label{app:exp setup}

\subsection{Hyper-parameters}
\label{app:exp hyper}

We follow the hyper-parameters described in Appendix B.2 of \citet{chen2024exploring} and perform 500 runs of hyper-parameter tuning using a Bayesian searcher for each method, reporting the best performance. We set the number of attention heads to 1 without searching this parameter. The hyper-parameters we searched are listed in Table \ref{tab:hyperparameter}. The parameters we used in our model are presented in Table \ref{tab:hyper finetune}, \ref{tab:hyper da}, and \ref{tab:hyper pt}.

\begin{table}[!h]
    \centering
    \begin{tabular}{l|c}
        \toprule
        \textbf{Hyper-parameters} & \textbf{Values}             \\ \midrule
        \texttt{Hidden Dimension} & \{8, 16, 32, 64, 128, 256\} \\
        \texttt{Number of Layers} & \{1, 2, 3\}                 \\
        \texttt{Normalize}        & \{none, batchnorm\}         \\
        \texttt{Learning Rate}    & \{5e-2, 1e-2, 5e-3, 1e-3\}  \\
        \texttt{Weight Decay}     & \{0.0, 5e-5, 1e-4, 5e-4\}   \\
        \texttt{Dropout}          & \{0.0, 0.1, 0.5, 0.8\}      \\
        \bottomrule
    \end{tabular}
    \vspace{-7pt}
    \caption{The hyper-parameters we searched.}
    \label{tab:hyperparameter}
\end{table}

\begin{table*}[!t]
    \resizebox{\linewidth}{!}{
        \begin{tabular}{llcccccc}
            \toprule
            Dataset & Graph Type & \# Dim & \# Layers & Normalize & Learning Rate & Weight Decay & Dropout \\ \midrule
            Cora    & Text-Rich  & 128    & 2         & None      & 1e-3          & 1e-4         & 0.8     \\
            Pubmed  & Text-Rich  & 256    & 2         & None      & 5e-3          & 0.0005       & 0.1     \\
            Cora    & Text-Limit & 256    & 3         & None      & 1e-3          & 5e-5         & 0.8     \\ \midrule
            Pubmed  & Text-Limit & 128    & 2         & None      & 1e-3          & 5e-5         & 0.5     \\
            USA     & Text-Free  & 128    & 3         & None      & 5e-2          & 1e-4         & 0.1     \\ \midrule
            Brazil  & Text-Free  & 8      & 3         & None      & 5e-3          & 1e-4         & 0.5     \\
            Europe  & Text-Free  & 256    & 2         & None      & 1e-3          & 5e-5         & 0.5     \\ \bottomrule
        \end{tabular}
    }
    \caption{The hyper-parameters for our TANS in basic training setting (i.e., training from scratch).}
    \label{tab:hyper finetune}
\end{table*}

\begin{table*}[!t]
    \resizebox{\linewidth}{!}{
        \begin{tabular}{llcccccc}
            \toprule
            Source Graph    & Target Graph    & Hidden Dim & Num Layers & Normalize & Learning Rate & Decay & Dropout \\ \midrule
            \texttt{USA}    & \texttt{Brazil} & 64         & 3          & None      & 1e-2          & 5e-5  & 0       \\
            \texttt{USA}    & \texttt{Europe} & 64         & 2          & Batch     & 5e-3          & 1e-4  & 0.5     \\ \midrule
            \texttt{Brazil} & \texttt{USA}    & 8          & 3          & Batch     & 1e-2          & 0     & 0.5     \\
            \texttt{Brazil} & \texttt{Europe} & 16         & 3          & None      & 5e-2          & 1e-4  & 0       \\ \midrule
            \texttt{Europe} & \texttt{USA}    & 16         & 2          & Batch     & 5e-3          & 1e-4  & 0.5     \\
            \texttt{Europe} & \texttt{Brazil} & 32         & 2          & Batch     & 5e-3          & 0     & 0.8     \\
            \bottomrule
        \end{tabular}
    }
    \caption{The hyper-parameters for our TANS in domain adaptation setting. }
    \label{tab:hyper da}

\end{table*}

\begin{table*}[!t]
    \resizebox{\linewidth}{!}{
        \begin{tabular}{llcccccc}
            \toprule
            Dataset                             & Graph Type & Hidden Dim & Num Layers & Normalize & Learning Rate & Decay & Dropout \\ \midrule
            \texttt{Cora} $\to$ \texttt{Pubmed} & Text-Rich  & 64         & 2          & None      & 5e-2          & 1e-4  & 0.5     \\
            \texttt{Pubmed} $\to$ \texttt{Cora} & Text-Rich  & 32         & 2          & None      & 1e-2          & 1e-4  & 0.1     \\ \midrule
            \texttt{Cora} $\to$ \texttt{Pubmed} & Text-Limit & 32         & 2          & None      & 5e-3          & 0     & 0.5     \\
            \texttt{Pubmed} $\to$ \texttt{Cora} & Text-Limit & 32         & 2          & None      & 5e-3          & 0     & 0.1     \\
            \bottomrule
        \end{tabular}
    }
    \caption{The hyper-parameters for our TANS in pretrain \& finetune setting. }
    \label{tab:hyper pt}
\end{table*}

\subsection{Pretrain \& Finetune Setting}
\label{app:exp pt}

In the transfer learning setting, the original node features of \texttt{Cora} (1,433 dimensions) and \texttt{Pubmed} (500 dimensions) cannot be directly used, as a single GNN cannot handle graphs with different feature dimensions. However, by using LLMs to encode the textual descriptions of nodes, we can naturally align the node features across different graphs. In this setting, we analyze the transfer learning performance on these datasets in the text-rich low-label scenario. The key difference from basic learning, where a model is trained from scratch, is that we use a pretrained model to initialize the parameters. During pretraining, we fix the number of epochs to 100 and the learning rate to 0.001.

\subsection{Domain Adaptation Setting}
\label{app:exp da}

Our proposed method converts basic graphs into text-attributed graphs, enabling the alignment of feature spaces across graphs. We evaluate our method in a classic cross-graph learning setting: domain adaptation. Domain adaptation transfers knowledge from a source graph to a target graph without fine-tuning, meaning the model is pretrained on the source graph and directly applied for inference on the target graph. In our experiments, we use the source graph for training, with 20\% of the nodes in the target graph randomly selected for validation and the remaining 80\% for testing. We evaluate our approach using three text-free airport graphs—\texttt{USA}, \texttt{Brazil}, and \texttt{Europe}—due to their aligned label spaces.

\section{Additional Experimental Results}

Additional experimental results are provided as follows. Table \ref{tab:text-free mlp} presents the results using the MLP encoder on text-free graphs. The complete ablation study results are shown in Table \ref{tab:ablation full}, and the full results for different textual encoders are presented in Table \ref{tab:text enc full}.

\begin{table*}[!t]
    \resizebox{\linewidth}{!}{
        \begin{tabular}{lcccccccc}
            \toprule
                                   & \multicolumn{4}{c}{\bf Low-Label} & \multicolumn{4}{c}{\bf High-Label}                                                                                                                                    \\ \cmidrule(lr){2-5}\cmidrule(lr){6-9}
                                   & \texttt{Europe}                   & \texttt{USA}                       & \texttt{Brazil}        & \textbf{Avg.}  & \texttt{Europe}       & \texttt{USA}          & \texttt{Brazil}       & \textbf{Avg.}  \\ \midrule
            Raw Feat. (One-Hot)    & 40.65 ± 6.30                      & 25.29 ± 1.36                       & 18.18 ± 1.30           & 28.04          & 25.31 ± 4.31          & 24.73 ± 1.96          & 23.58 ± 6.66          & 24.54          \\ \midrule
            Node Degree            & {46.22 ± 3.65}                    & {36.08 ± 2.09}                     & {51.03 ± 14.49}        & {44.44}        & {45.64 ± 4.01}        & \textbf{49.08 ± 3.18} & 32.84 ± 11.66         & {42.52}        \\
            Eigenvector            & 44.56 ± 5.07                      & 30.17 ± 1.88                       & 40.91 ± 12.80          & 38.55          & 33.62 ± 7.48          & 33.59 ± 2.96          & 22.22 ± 7.65          & 29.81          \\
            Random Walk            & 44.49 ± 3.16                      & 25.00 ± 0.09                       & 49.39 ± 14.03          & 39.63          & 21.23 ± 3.88          & 23.70 ± 2.41          & {45.93 ± 8.41}        & 30.29          \\ \midrule
            \rowcolor{Gray}   TANS & \textbf{46.77 ± 3.18}             & \textbf{40.35 ± 2.26}              & \textbf{52.42 ± 17.04} & \textbf{46.51} & \textbf{47.33 ± 4.80} & {46.25 ± 2.76}        & \textbf{48.52 ± 7.43} & \textbf{47.37} \\             \bottomrule
        \end{tabular}
    }
    \vspace{-7pt}
    \caption{The results on text-free airport graphs with MLP backbone. }
    \label{tab:text-free mlp}
\end{table*}

\begin{table*}[!t]
    \centering
    \resizebox{0.8\linewidth}{!}{
    \begin{tabular}{llcccc}
        \toprule
                                          &                             & \textbf{GCN}          & \textbf{GAT}          & \textbf{MLP}          & \textbf{Avg.}  \\ \midrule
        \multirow{3}{*}{\bf Text-Rich}    & Raw Text                    & 79.19 ± 1.63          & 80.09 ± 1.57          & 70.55 ± 1.40          & 76.61          \\ \cmidrule{2-6}
                                          & \quad + TANS w.o. neighbors & \textbf{81.53 ± 1.47} & 80.71 ± 1.21          & 70.37 ± 1.96          & 77.54          \\
                                          & \quad + TANS                & 81.26 ± 1.48          & \textbf{81.08 ± 1.62} & \textbf{72.47 ± 1.96} & \textbf{78.27} \\ \midrule\midrule
        \multirow{3}{*}{\bf Text-Limited} & Raw Text                    & 79.06 ± 1.68          & 77.89 ± 1.64          & 57.42 ± 2.46          & 71.46          \\ \cmidrule{2-6}
                                          & \quad + TANS w.o. neighbors & 79.32 ± 1.61          & 78.64 ± 1.55          & 60.80 ± 2.05          & 72.92          \\
                                          & \quad + TANS                & \textbf{79.94 ± 1.62} & \textbf{80.34 ± 1.25} & \textbf{68.35 ± 1.85} & \textbf{76.21} \\
        \bottomrule
    \end{tabular}
    }
    \caption{The full ablation study results without providing neighborhood information in generating node descriptions. We use \texttt{Cora} in low-label setting.}
    \label{tab:ablation full}
\end{table*}

\begin{table*}[!t]
    \centering
    \resizebox{\linewidth}{!}{
        \begin{tabular}{lccccc}
            \toprule
                                           & \textbf{MiniLM} \citep{wang2020minilm} & \textbf{Albert} \citep{Lan2020ALBERT} & \textbf{Roberta} \citep{liu2020roberta} & \textbf{MPNet} \citep{song2020mpnet} & \textbf{Avg.}  \\ \midrule
            Raw Text                       & 79.19 ± 1.63                           & 77.76 ± 1.45                          & 81.43 ± 1.42                            & 82.04 ± 1.49                         & 80.11          \\
            \quad + TAPE                   & 79.64 ± 1.36                           & 77.88 ± 1.41                          & 81.78   ± 1.55                          & 82.70 ± 1.06                         & 80.50          \\
            \quad + KEA                    & 80.08 ± 1.71                           & 78.23 ± 1.48                          & 81.75 ± 1.49                            & 82.23 ± 1.28                         & 80.57          \\ \midrule
            \rowcolor{Gray}   \quad + TANS & \textbf{81.26 ± 1.48}                  & \textbf{78.46 ± 1.19}                 & \textbf{82.29 ± 1.65}                   & \textbf{82.88 ± 1.48}                & \textbf{81.22} \\
            \bottomrule
        \end{tabular}
    }
    \caption{Full results of different textual encoder on text-rich \texttt{Cora} with low-label setting and GCN backbone.}
    \label{tab:text enc full}
\end{table*}

\section{Additional Ablation Studies}

\subsection{Advanced Encoder}

We also compare our methods to TAPE and KEA on an advanced graph encoder, OFA \citep{liu2024one}. OFA proposes a graph foundation model that leverages LLMs to align node features across graphs. This model does not generate additional textual information. Instead, it employs a template-based method to help language models better encode the original node text (e.g., "Feature node. Node title: <paper title>, node abstract: <paper abstract>, …"). We consider OFA as a "backbone" model, similar to GCN, as it can serve as the base encoder for TAG-based methods. We conducted additional experiments on the low-label Cora dataset (text-rich graph), using OFA as the backbone, as shown in Table \ref{tab:ofa results}. It is worth noting that OFA's performance is lower than GCN's, which is likely because OFA is designed for handling cross-domain and cross-task graphs, whereas GCN is optimized for solving single tasks individually.

\begin{table}[!h]
    \centering
    \resizebox{\linewidth}{!}{
    \begin{tabular}{lcccc}
        \toprule
    \textbf{Methods} &                        Raw Text                   & TAPE & KEA & TANS \\ \midrule
    \textbf{Accuracy} &  76.61 ± 3.28               & 77.85 ± 2.14    & 78.01 ± 2.59   & \textbf{78.53 ± 1.61}    \\ \bottomrule
    \end{tabular}
    }
    \caption{The results of using OFA \citep{liu2024one} as advanced encoder on low-label text-rich Cora.}
    \label{tab:ofa results}
\end{table}

\subsection{Impacts of Topological Information on Prompt Design}

We provide additional experimental results to demonstrate how incorporating topological properties enhances model performance. In particular, we simply remove the corresponding prompt to analyze the impact of topologies. Using the Cora dataset in the low-label setting, we evaluate both text-rich and text-limited scenarios. As shown in Table \ref{tab:topo prompt impact}, removing topological information leads to a drop in performance. Regarding the relative importance of different topological properties, we consider their significance may vary depending on the dataset and graph type. For instance, in social networks, properties like clustering coefficient may be more important as they capture triangular patterns that indicate strong friendship relationships. Similarly, other graphs may favor different topological patterns. To account for this variability, we aim to provide a comprehensive set of topological features, allowing the LLMs to automatically identify and leverage the patterns that are most beneficial for the specific dataset and task.

\begin{table}[!h]
    \centering
    \resizebox{\linewidth}{!}{
    \begin{tabular}{lccc}
        \toprule
               & Raw Text     & + w.o. Topo &  + TANS \\ \midrule
    \textbf{Text-Rich} & 79.19 ± 1.63 & 80.69 ± 2.63              & \textbf{81.26 ± 1.48}    \\
    \textbf{Text-Limit} & 79.06 ± 1.68 & 79.45 ± 1.66              & \textbf{79.94 ± 1.62}    \\
    \bottomrule
    \end{tabular}
    }
    \caption{The impact of topological information, i.e., node properties, on the prompt design. The results are based on low-label Cora. }
    \label{tab:topo prompt impact}
\end{table}

\subsection{Impacts of Topological Information on Node Features}

It is possible to directly use topological properties, such as degree, centrality, and clustering coefficients, as node features (or append them to the original node features). However, this approach is limited to single-graph training and cannot be effectively extended to cross-graph training unless the original node features are either aligned or excluded entirely.

We conducted additional experiments to compare the effectiveness of using LLM-generated node descriptions versus directly using topological properties as features. These experiments include (1) single-graph training with a low-label setting using a GCN backbone (Table \ref{tab:tpf single}) and (2) cross-graph training (domain adaptation) with a low-label setting using the same backbone (Table \ref{tab:tpf cross}). We established a baseline, \textbf{Topology Properties as Features (TPF)}, where the topological properties used in prompt design were concatenated as node features. These features underwent normalization for training stability. 

\begin{table}[!h]
    \centering
    \resizebox{\linewidth}{!}{
    \begin{tabular}{lccc}
        \toprule
                                      & \texttt{Europe}       & \texttt{USA}          & \texttt{Brazil}        \\ \midrule
    TPF & 54.90 ± 4.68 & 59.66 ± 3.01 & 72.48 ± 14.34 \\
    TANS                              & \textbf{56.87 ± 3.14} & \textbf{61.08 ± 2.71} & \textbf{80.61 ± 12.14} \\
    \bottomrule
    \end{tabular}
    }
    \caption{Comparison between TANS and Topology Properties as Features (TPF) on single-graph learning. }
    \label{tab:tpf single}
\end{table}

\begin{table}[!h]
    \centering
    \resizebox{\linewidth}{!}{
    \begin{tabular}{lcccc}
        \toprule
         & \texttt{USA}$\to$\texttt{Europe} & \texttt{USA}$\to$\texttt{Brazil} & \texttt{Brazil}$\to$\texttt{USA} & \texttt{Brazil}$\to$\texttt{Europe} \\ \midrule
    TPF  & 47.51 ± 3.20             & 54.12 ± 3.25             & 50.38 ± 3.71             & 50.88 ± 4.63                \\
    TANS & 50.99 ± 3.31             & 67.17 ± 4.68             & 54.96 ± 1.80             & 53.79 ± 2.15                \\
     \bottomrule
    \end{tabular}
    }
    \caption{Comparison between TANS and Topology Properties as Features (TPF) on cross-graph learning. }
    \label{tab:tpf cross}
\end{table}

The results show that TANS significantly outperforms the baseline in both single-graph and cross-graph training. This improvement may be because the raw topological properties are too comprehensive for GNNs to effectively discern the most relevant features or correlations among them. In contrast, LLMs appear to infer potential relationships between these properties more effectively, enabling TANS to achieve superior performance. 

\subsection{Advanced Knowledge Transfer}

We conduct experiments on transfer learning across text-rich and text-free datasets, as well as across different domains (e.g., citation networks and airport networks). This experiments may better demonstrate the potential of TANS in jointly handling text-attributed and text-free graphs. Note that citation networks (Cora, Pubmed) and airport networks (USA, Europe) differ significantly in their underlying structures and semantics, which naturally leads to lower transfer performance across domains. However, we observe that the performance drop is marginal in some cases, likely due to shared structural patterns such as the importance of high-degree nodes. The results are presented in Table \ref{tab:advanced transfer}. These results demonstrate that TANS can support transfer learning across both similar and dissimilar domains, achieving reasonable performance even when transferring between fundamentally different types of graphs.

\begin{table}[!h]
    \centering
    \resizebox{\linewidth}{!}{
    \begin{tabular}{llc}
        \toprule
    \textbf{Source}           & \textbf{Target}         & \textbf{Accuracy}     \\ \midrule
    Text-rich pubmed & Text-rich Cora & 80.05 ± 1.74 \\
    Text-free USA    & Text-rich Cora & 79.53 ± 2.51 \\ \midrule
    Text-free Europe & Text-free USA  & 56.31 ± 1.24 \\
    Text-rich Cora   & Text-free USA  & 55.49 ± 2.59 \\
    \bottomrule
    \end{tabular}
    }
    \caption{The results of transfer learning between text-attributed and text-free graphs.}
    \label{tab:advanced transfer}
\end{table}

\section{Results on Larger Graphs}

Following \citep{he2024harnessing}, we conduct experiments on ogbn-products dataset to evaluate the model performance on relatively larger graphs. Note that we follow \citep{he2024harnessing} to conduct subgraph sampling to manage computational costs. Specifically, for the ogbn-products dataset ($\sim$2,500,000 nodes), we sample a smaller graph with 54,000 nodes and conduct experiments on this reduced graph. Note that we did not provide results for KEA as no processed text data was available. As shown in Table \ref{tab:products results}, our proposed method, TANS, consistently outperforms TAPE on this relatively large graph. This can be attributed to TANS's ability to leverage topological information in graphs by utilizing LLMs for inference. We aim to include experiments on large-scale graphs, e.g., ogbn-arxiv ($\sim$150,000 nodes) in our future work.

\begin{table}[!h]
    \centering
    \resizebox{\linewidth}{!}{
    \begin{tabular}{lccc}
        \toprule
     & \textbf{MLP}          & \textbf{GCN}          & \textbf{GAT} \\ \midrule
    Original Text & 53.85 ± 0.17	& 70.52 ± 0.51	& - \\
    \quad + TAPE          & 73.65 ± 0.60 & 77.49 ± 0.54 & 77.64 ± 0.58            \\
    \rowcolor{Gray} \quad + TANS          & 74.30 ± 0.85 & 78.05 ± 0.33 & 78.35 ± 0.35      \\     
    \bottomrule
    \end{tabular}
    }
    \caption{Model Performance on ogbn-products.}
    \label{tab:products results}
\end{table}

\section{Case Studies}

We present case studies in the following pages. Our findings show that incorporating topological information significantly influences the generated answers and improves the quality of the generated texts. Furthermore, providing neighborhood information allows the LLMs to adjust their predictions, leading to more accurate results. This highlights the importance of leveraging both node-specific and neighborhood data for improved performance in using LLMs to synthesize node descriptions.

\begin{table*}[!t]
    \centering
    \resizebox{\linewidth}{!}{
        \begin{tabular}{p{2.1cm} p{15cm}}
            \toprule
            \rowcolor{Gray}   \multicolumn{2}{l}{\textit{{Case 1: Text-Rich Graph -- Cora}}}                                                                                                                                                                                                                                                                                                                                                                                                                                                                                                                                                                                                                                                                                                                                                                                                                                                                                                                                                                                                                                                                                                                                                                                                                     \\ \midrule
            \textbf{Node Idx}                                         & 2                                                                                                                                                                                                                                                                                                                                                                                                                                                                                                                                                                                                                                                                                                                                                                                                                                                                                                                                                                                                                                                                                                                                                                                                                                        \\
            \textbf{Node Class}                                       & Case Based                                                                                                                                                                                                                                                                                                                                                                                                                                                                                                                                                                                                                                                                                                                                                                                                                                                                                                                                                                                                                                                                                                                                                                                                                               \\
            \textbf{Win}                                              & Generated Text with Full Prompt                                                                                                                                                                                                                                                                                                                                                                                                                                                                                                                                                                                                                                                                                                                                                                                                                                                                                                                                                                                                                                                                                                                                                                                                          \\
            \midrule
            \textbf{Generated Text with Full Prompt}                  & \texttt{"Based on the provided information, the node associated with ""Design by Interactive Exploration Using Memory-Based Techniques"" can be classified into the following three potential classes: 1. \textbf{**Case Based**}: The paper focuses on utilizing memory-based techniques, which aligns closely with case-based reasoning. It emphasizes the importance of past experiences in design, suggesting that it fits well within this category. 2. \textbf{**Theory**}: The concepts discussed in the paper pertain to theoretical foundations of design exploration and memory-based techniques in artificial intelligence, making it relevant to the Theory class. It discusses frameworks for understanding design processes and knowledge representation. 3. \textbf{**Probabilistic Methods**}: Although not explicitly stated, the use of memory-based techniques often involves probabilistic reasoning to retrieve and adapt past design cases to current contexts, indicating relevance in this area as well. The other classes, such as Genetic Algorithms or Neural Networks, seem less applicable since the focus is primarily on memory-based design techniques rather than evolutionary or neural approaches." } \\ \midrule
            \textbf{Generated Text with Prompt without Neighborhoods} & \texttt{Based on the node's description and properties, the three potential classes for this citation node are: 1. \textbf{**Rule Learning**}: The node discusses an algorithm for feature engineering driven by first-order clauses, which aligns closely with methodologies used in rule learning where rules or patterns are extracted from data. 2. \textbf{**Probabilistic Methods**}: The stochastic nature of the proposed algorithm indicates a probabilistic approach to feature selection, making it relevant to the field of probabilistic methods, which utilize uncertainty in learning. 3. \textbf{**Theory**}: The node provides insights into a well-established aspect of propositional learning and the necessity of good features, suggesting a theoretical contribution to the understanding of learning algorithms and their performance. These classes resonate with the search for optimal representations in machine learning and acknowledge both the technical algorithmic contributions and theoretical implications outlined in the node.}                                                                                                                                                                   \\

            \bottomrule
        \end{tabular}
    }
\end{table*}

\begin{table*}[!t]
    \centering
    \resizebox{\linewidth}{!}{
        \begin{tabular}{p{2.1cm} p{15cm}}
            \toprule
            \rowcolor{Gray}   \multicolumn{2}{l}{\textit{{Case 1: Text-Rich Graph -- Cora}}}                                                                                                                                                                                                                                                                                                                                                                                                                                                                                                                                                                                                                                                                                                                                                                                                                                                                                                                                                                                                                                                                                                                                                                                                                                                 \\ \midrule
            \textbf{Node Idx}                                         & 1199                                                                                                                                                                                                                                                                                                                                                                                                                                                                                                                                                                                                                                                                                                                                                                                                                                                                                                                                                                                                                                                                                                                                                                                                                                                                 \\
            \textbf{Node Class}                                       & Genetic Algorithms                                                                                                                                                                                                                                                                                                                                                                                                                                                                                                                                                                                                                                                                                                                                                                                                                                                                                                                                                                                                                                                                                                                                                                                                                                                   \\
            \textbf{Win}                                              & Generated Text with Full Prompt                                                                                                                                                                                                                                                                                                                                                                                                                                                                                                                                                                                                                                                                                                                                                                                                                                                                                                                                                                                                                                                                                                                                                                                                                                      \\
            \midrule
            \textbf{Generated Text with Full Prompt}                  & \texttt{Based on the provided information about the node titled ""How to evolve autonomous robots,"" three potential classes for the node are: 1. \textbf{**Genetic Algorithms**}: The node discusses evolutionary strategies for robots, indicating a focus on optimization and adaptation, core elements of genetic algorithms as evidenced by the connectivity with similar nodes that explore genetic algorithms in neuro-evolution. 2. \textbf{**Neural Networks**}: Given that several similar nodes involve artificial neural networks used for controlling robots, it is reasonable to classify this node within the neural networks class, as neural architectures are often evolved using methods discussed in the context of autonomous robots. 3. \textbf{**Theory**}: The context of the paper hints at theoretical foundations for evolving autonomous systems, emphasizing principles and methodologies relevant to evolutionary approaches in robotics, which aligns with the general theoretical framework underlying both genetic algorithms and neural networks. These classifications are supported by the content of the similar nodes that emphasize evolution, neural architectures, and fundamental principles guiding these methodologies.} \\ \midrule
            \textbf{Generated Text with Prompt without Neighborhoods} & \texttt{Based on the provided characteristics of the paper node, the potential classes are: 1. \textbf{**Reinforcement Learning**}: Given that the title ""How to evolve autonomous robots"" suggests a focus on enabling machines to learn from interactions with their environment, this aligns with core principles of reinforcement learning. 2. \textbf{**Case Based**}: The mention of evolution implies a process of learning from previous cases, which is a hallmark of case-based reasoning methods where past experiences inform decision-making strategies in novel situations. 3. \textbf{**Neural Networks**}: Although less directly implied by the title, autonomous robots often utilize neural networks for perception and decision-making processes, making this a reasonable classification given the context of evolving robotic intelligence. These classes are assessed based on the paper's focus on autonomy and evolution in robotics, which are fundamental themes in reinforcement learning and case-based approaches, while neural networks are often employed to support such advancements.}                                                                                                                                           \\

            \bottomrule
        \end{tabular}
    }
\end{table*}

\begin{table*}[!t]
    \centering
    \resizebox{\linewidth}{!}{
        \begin{tabular}{p{2.1cm} p{15cm}}
            \toprule
            \rowcolor{Gray}   \multicolumn{2}{l}{\textit{{Case 1: Text-Rich Graph -- Cora (Failure Case)}}}                                                                                                                                                                                                                                                                                                                                                                                                                                                                                                                                                                                                                                                                                                                                                                                                                                                                                                                                                                                                                                                                                                                                                                                                                       \\ \midrule
            \textbf{Node Idx}                                         & 523                                                                                                                                                                                                                                                                                                                                                                                                                                                                                                                                                                                                                                                                                                                                                                                                                                                                                                                                                                                                                                                                                                                                                                                                                                                       \\
            \textbf{Node Class}                                       & Rule Learning                                                                                                                                                                                                                                                                                                                                                                                                                                                                                                                                                                                                                                                                                                                                                                                                                                                                                                                                                                                                                                                                                                                                                                                                                                             \\
            \textbf{Win}                                              & Tie                                                                                                                                                                                                                                                                                                                                                                                                                                                                                                                                                                                                                                                                                                                                                                                                                                                                                                                                                                                                                                                                                                                                                                                                                                                       \\
            \midrule
            \textbf{Generated Text with Full Prompt}                  & \texttt{"Based on the information provided about the paper ""Drug design by machine learning: Modelling drug activity,"" the potential classes for this node are: 1. \textbf{**Probabilistic Methods**}: The paper discusses the application of machine learning systems, which often utilize probabilistic techniques to model and predict outcomes in complex datasets, aligning well with this category. 2. \textbf{**Neural Networks**}: Given the mention of machine learning tools, it's likely that neural networks might be among the techniques applied, especially for problems in drug activity modeling. 3. \textbf{**Genetic Algorithms**}: While this is a more indirect connection, the reference to optimizing outcomes in drug design suggests that evolutionary approaches like genetic algorithms might be relevant in conjunction with machine learning methods. The attributes such as low clustering coefficients and high node degree imply the node's connections are sparse yet significant, indicating a central role in its niche, possibly aligning with exploratory frameworks in machine learning, particularly in the probabilistic and neural network domains."}                                                          \\ \midrule
            \textbf{Generated Text with Prompt without Neighborhoods} & \texttt{"Based on the description of the paper ""Drug design by machine learning: Modelling drug activity,"" the potential classes for the node can be evaluated as follows: 1. \textbf{**Neural Networks**}: Since the paper discusses machine learning tools and the performance enhancement they provide over traditional methods, it implies the use of advanced machine learning techniques, which likely include neural networks as a popular method in drug design and QSAR modeling. 2. \textbf{**Genetic Algorithms**}: The context of optimizing drug activity modeling may involve mechanisms akin to genetic algorithms, which are often used for optimizing complex search and solution spaces, although this is less explicit from the content provided. 3. \textbf{**Theory**}: The foundational principles around drug activity modeling, especially in relation to the Hansch method and machine learning systems, categorize this work under theoretical discussions in a computational context.  In summary, the key themes of machine learning applications in drug design suggest a strong connection to Neural Networks, with possible ties to Genetic Algorithms and theoretical explorations in machine learning methodologies."} \\

            \bottomrule
        \end{tabular}
    }
\end{table*}

\begin{table*}[!t]
    \centering
    \resizebox{\linewidth}{!}{
        \begin{tabular}{p{2.1cm} p{15cm}}
            \toprule
            \rowcolor{Gray}   \multicolumn{2}{l}{\textit{{Case 2: Text-Limit Graph -- Cora}}}                                                                                                                                                                                                                                                                                                                                                                                                                                                                                                                                                                                                                                                                                                                                                                                                                                                                                                                                                                                                                                                                                                              \\ \midrule
            \textbf{Node Idx}                                         & 525                                                                                                                                                                                                                                                                                                                                                                                                                                                                                                                                                                                                                                                                                                                                                                                                                                                                                                                                                                                                                                                                                                                                \\
            \textbf{Node Class}                                       & Neural Network                                                                                                                                                                                                                                                                                                                                                                                                                                                                                                                                                                                                                                                                                                                                                                                                                                                                                                                                                                                                                                                                                                                     \\
            \textbf{Win}                                              & Generated Text with Full Prompt                                                                                                                                                                                                                                                                                                                                                                                                                                                                                                                                                                                                                                                                                                                                                                                                                                                                                                                                                                                                                                                                                                    \\
            \midrule
            \textbf{Generated Text with Full Prompt}                  & \texttt{
                "Based on the information provided and the context of the citation network graph, I would classify the node into the following three potential classes: 1. **Theory**: Given the node's description as a paper in molecular, cellular, and developmental biology, along with the lack of a specific application or algorithm, the content likely discusses foundational theories relevant to genetic research and methodologies. 2. \textbf{**Genetic Algorithms**}: Although not explicitly stated, the mention of ""Identification of Protein Coding Regions"" suggests a focus on genomic patterns, which may employ genetic algorithms for optimization and analysis in bioinformatics. 3. \textbf{**Neural Networks**}: The similar nodes, particularly the one discussing ""theory of neural computation,"" imply a potential relationship with neural network methodologies, which are often employed in biological data analysis, including gene sequence classification. These classifications reflect the theoretical and applied aspects of the research paper, as indicated by its connections to relevant topics in computational biology."
            }                                                                                                                                                                                                                                                                                                                                                                                                                                                                                                                                                                                                                                                                                                                                                                                                                                                                                                                                                                                                                                                                                                                                                                                              \\ \midrule
            \textbf{Generated Text with Prompt without Neighborhoods} & \texttt{
            Based on the given information about the node, the most suitable potential classes are **'Theory', 'Genetic Algorithms', and 'Case Based'**.  1. \textbf{**Theory**}: The node represents a paper focused on identifying protein-coding regions, which indicates a strong foundation in biological and computational theories regarding gene identification and genomic analysis. 2. \textbf{**Genetic Algorithms**}: The paper's title and keywords suggest it may involve computational methods akin to genetic algorithms, which are often used in genomics and bioinformatics to optimize solutions and models involving genetic data. 3. \textbf{**Case Based**}: The research might involve case-based methods, especially as it pertains to comparing genomic sequences or models based on existing cases of gene annotation or identification. The low values in clustering and node degree indicate that this paper may be less cited and not central in the network, aligning with a theoretical perspective rather than a method-driven or application-focused approach often seen in more active research domains like Neural Networks or Reinforcement Learning.                } \\

            \bottomrule
        \end{tabular}
    }
\end{table*}

\begin{table*}[!t]
    \centering
    \resizebox{\linewidth}{!}{
        \begin{tabular}{p{2.1cm} p{15cm}}
            \toprule
            \rowcolor{Gray}   \multicolumn{2}{l}{\textit{{Case 2: Text-Limit Graph -- Cora}}}                                                                                                                                                                                                                                                                                                                                                                                                                                                                                                                                                                                                                                                                                                                                                                                                                                                                                                                                                                                                                                                                                                                                                                                                 \\ \midrule
            \textbf{Node Idx}                                         & 1936                                                                                                                                                                                                                                                                                                                                                                                                                                                                                                                                                                                                                                                                                                                                                                                                                                                                                                                                                                                                                                                                                                                                                                                                                  \\
            \textbf{Node Class}                                       & Neural Network                                                                                                                                                                                                                                                                                                                                                                                                                                                                                                                                                                                                                                                                                                                                                                                                                                                                                                                                                                                                                                                                                                                                                                                                        \\
            \textbf{Win}                                              & Generated Text with Full Prompt                                                                                                                                                                                                                                                                                                                                                                                                                                                                                                                                                                                                                                                                                                                                                                                                                                                                                                                                                                                                                                                                                                                                                                                       \\
            \midrule
            \textbf{Generated Text with Full Prompt}                  & \texttt{
            Based on the node's title, ""Clustering learning tasks and the selective cross-task transfer of knowledge,"" and its connectivity information, the following potential classes can be assessed: 1. \textbf{**Case Based**}: The focus on clustering and transfer of knowledge suggests that the node might be related to case-based reasoning, where previous experiences inform new problem-solving. 2. \textbf{**Neural Networks**}: Given the emphasis on learning tasks, particularly in complex environments, neural networks could relate through their application in clustering methodologies and knowledge transfer. 3. \textbf{**Theory**}: The title implies a theoretical exploration of learning tasks. A theoretical class encompasses foundational studies in learning, making it a suitable classification given the lack of direct application or methodology in the description. These classes are derived from the node's focus on learning methodologies and abstract concepts, supported by the similarity to other titles which also deal with learning and knowledge transfer. The low clustering and centrality metrics indicate that this node does not connect strongly with others, highlighting a potential niche theoretical focus.                } \\ \midrule
            \textbf{Generated Text with Prompt without Neighborhoods} & \texttt{
                Given the properties of the node, the potential classes it could belong to are: 1. \textbf{**Rule Learning**}: The low clustering coefficients and degree suggest that the paper may focus on specific aspects of knowledge transfer and clustering tasks without much interconnectivity with other research topics, which aligns well with rule-based approaches that aim for precision in knowledge representation. 2. \textbf{**Theory**}: The lack of citations and low degree implies that this paper might be theoretical in nature, possibly proposing a new framework or model for clustering learning tasks. The theoretical foundations often precede practical implementations and might explain the sparse connections. 3. \textbf{**Probabilistic Methods**}: The content related to selective cross-task transfer of knowledge may imply a consideration of uncertainty and inference, which is common in probabilistic approaches to machine learning. This could also relate to knowledge representation and classification tasks. The other classes like Neural Networks or Genetic Algorithms seem less likely given the citation network's low engagement for this node.
            }                                                                                                                                                                                                                                                                                                                                                                                                                                                                                                                                                                                                                                                                                                                                                                                                                                                                                                                                                                                                                                                                                                                                                                                                                                                                                 \\

            \bottomrule
        \end{tabular}
    }
\end{table*}

\begin{table*}[!t]
    \centering
    \resizebox{\linewidth}{!}{
        \begin{tabular}{p{2.1cm} p{15cm}}
            \toprule
            \rowcolor{Gray}   \multicolumn{2}{l}{\textit{{Case 3: Text-Free Graph -- USA}}}                                                                                                                                                                                                                                                                                                                                                                                                                                                                                                                                                                                                                                                                                                                                                                                                                                                                                                                                                                                                                                                                                                                                                                                                           \\ \midrule
            \textbf{Node Idx}                        & 590                                                                                                                                                                                                                                                                                                                                                                                                                                                                                                                                                                                                                                                                                                                                                                                                                                                                                                                                                                                                                                                                                                                                                                                                                                            \\
            \textbf{Node Class}                      & High Activity                                                                                                                                                                                                                                                                                                                                                                                                                                                                                                                                                                                                                                                                                                                                                                                                                                                                                                                                                                                                                                                                                                                                                                                                                                  \\\midrule
            \textbf{Generated Text with Full Prompt} & \texttt{590	Based on the given properties of the airport node, it can be assessed that the node exhibits a high activity level within the airport network. 1. **Node Degree**: The node ranks 5th in node degree with a value of 0.1749, indicating it has a relatively high number of direct connections to other airports. This suggests a high level of connectivity and activity. 2. **Closeness Centrality**: Ranked 12th with a value of 0.4803, this property signifies that the node can reach other airports quickly, enhancing its operational effectiveness within the network. 3. **Betweenness Centrality**: With a rank of 15 and a value of 0.0256, the node plays a crucial role in connecting different parts of the network, further indicating its active role in facilitating connections. 4. **Clustering Coefficient**: The clustering coefficient (0.2882) and square clustering coefficient (0.2130) suggest a reasonable tendency to form tightly knit groups with other airports, which could enhance collaborative operational activities. Overall, the combination of these high rankings in degree, closeness, and betweenness centralities underline the node's importance and activity in the airport network.} \\ \midrule\midrule
            \textbf{Node Idx}                        & 822                                                                                                                                                                                                                                                                                                                                                                                                                                                                                                                                                                                                                                                                                                                                                                                                                                                                                                                                                                                                                                                                                                                                                                                                                                            \\
            \textbf{Node Class}                      & Low Activity                                                                                                                                                                                                                                                                                                                                                                                                                                                                                                                                                                                                                                                                                                                                                                                                                                                                                                                                                                                                                                                                                                                                                                                                                                   \\\midrule
            \textbf{Generated Text with Full Prompt} & \texttt{The activity level of the airport node can be assessed as low. This assessment is based on several key metrics: 1. **Node Degree**: The degree of 0.0025 and ranking of 142 indicates that the airport has relatively few direct connections to other airports compared to the rest of the network. A low degree suggests limited connectivity and hence lower activity. 2. **Betweenness Centrality**: With a betweenness centrality of 0.0000 and a ranking of 794, this node does not serve as a significant intermediary for traffic between other airports. This implies the airport is not a critical hub for travel, further indicating low activity. 3. **Square Clustering Coefficient**: The value of 0.0842 and ranking of 795 suggest that even when connected, the airport's connections do not create a highly interconnected cluster, which often indicates limited operational activity or network significance. Despite a perfect clustering coefficient of 1.0000, which may suggest strong local connectivity, the overall low rank and scores in degree and betweenness centrality strongly suggest that the airport operates on a low activity level within the broader network structure.}                       \\ \midrule\midrule
            \textbf{Node Idx}                        & 527                                                                                                                                                                                                                                                                                                                                                                                                                                                                                                                                                                                                                                                                                                                                                                                                                                                                                                                                                                                                                                                                                                                                                                                                                                            \\
            \textbf{Node Class}                      & Moderately Low Activity                                                                                                                                                                                                                                                                                                                                                                                                                                                                                                                                                                                                                                                                                                                                                                                                                                                                                                                                                                                                                                                                                                                                                                                                                        \\\midrule
            \textbf{Generated Text with Full Prompt} & \texttt{The airport node's activity level can be classified as moderate based on its properties. The **Node Degree** of 0.0059, which ranks 138th among 1190 nodes, indicates a relatively low number of direct connections to other airports, suggesting it is not a highly connected hub.  However, the **Clustering Coefficient** of 0.1905 (ranked 449) signifies a moderate tendency for the node to cluster with other connected nodes, which can enhance its connectivity within local groups. The **Closeness Centrality** at 0.3375 (ranked 378) suggests that while the node may not have a high degree, it can still reach other airports relatively efficiently, indicating a fair level of influence within the network. Lastly, the **Betweenness Centrality** of 0.0022 (ranked 142) implies that the node plays a limited role as a bridge between other more interconnected nodes, which could further support the assessment of moderate activity. In summary, while it is not a central or highly connected node, its clustering and closeness centralities indicate a reasonable level of activity in the airport network.}                                                                                                \\
            \bottomrule
        \end{tabular}
    }
\end{table*}

\end{document}